# FUZZY APPROACHES TO CONTEXT VARIABLES IN FUZZY GEOGRAPHICALLY WEIGHTED CLUSTERING

Nguyen Van Minh [1] and Le Hoang Son[2]

[1] Faculty of Science, Hanoi University of Natural Resources and Environment,
41A Phu Dien, Bac Tu Liem, Hanoi, Vietnam
`nvminh@hunre.edu.vn`

[2] VNU University of Science, Vietnam National University,
334 Nguyen Trai, Thanh Xuan, Hanoi, Vietnam
`sonlh@vnu.edu.vn`

*ABSTRACT*

*Fuzzy Geographically Weighted Clustering (FGWC) is considered as a suitable tool for the analysis of geo-demographic data that assists the provision and planning of products and services to local people. Context variables were attached to FGWC in order to accelerate the computing speed of the algorithm and to focus the results on the domain of interests. Nonetheless, the determination of exact, crisp values of the context variable is a hard task. In this paper, we propose two novel methods using fuzzy approaches for that determination. A numerical example is given to illustrate the uses of the proposed methods.*

## 1. INTRODUCTION

*Geo-Demographic Analysis* (GDA) is a major concentration of various interdisciplinary researches nowadays and has been being used in many decision–making processes involving the provision and distribution of products and services to communities. Results of GDA are often visualized on a map as several distinct groups that represent for different levels of a population's characteristic, e.g. "High density of chain-smokers" and "Low density of chain-smokers". Thus, they assist effectively for many decision–making processes involving the provision and distribution of products and services to communities, the determination of common population's characteristics and the study of population variation in terms of gender, ages, sex, ethnicity, etc. According to a review of typical examples of GDA in [11], GDA was proven to be one of the most promising researches in the scientific world nowadays. Some of the first methods applied to GDA are *Principal Component Analysis* (PCA) in [20] and *Self-Organizing Maps* (SOM) in [6] that rely one the basic principles of statistics and neural networks to determine the underlying demographic and socio-economic phenomena. However, the disadvantages of those methods are the requisition of large memory space and computational complexity. Indeed, clustering algorithms were opted instead. Two typical hard clustering methods used for GDA available in the literature are *Agglomerative Hierarchical Clustering* in [2] and *K-Means* in [9]. These algorithms classify geo-demographic datasets into clusters represented in forms of hierarchical trees and isolated groups. Data points in each group have similar ethnic and socio-economic characteristics. Nonetheless, using hard clustering for GDA often leads to the issues of ecological fallacy, which can be shortly understood that statistics accurately describing group characteristics do not necessarily apply to individuals within that group. For this fact, *Fuzzy C-Means* (FCM) and its variants were considered as the appropriate methods to determine the distribution of a demographic feature on a map as described in some articles such as [1, 5, 10, 12-19]. Since the results of FCM are independent to the geographical factors, some improvements of that algorithm were made by attaching FCM with a spatial model such as SIM in [3] and SIM-PF in [7, 16, 18]. The Fuzzy Geographically Weighted Clustering (FGWC) in [7] incorporated with SIM-PF is an effective algorithm for GDA. Nonetheless, the computing speed of FGWC is slow since the cluster membership modification process has to be done in each step. The authors in [18] introduced CFGWC using the context variable term to narrow the original dataset under some conditions of certain dimensions. Because only a subset of original dataset which has considerable meaning to the context is invoked, the velocity and efficiency of clustering can be improved considerably and the result focuses on the area that really has many relevant points. For example, if we want to look for a shopping area then a new context "shopping" will be put to the algorithm to reduce the search space. In case of little context variables, the speed of CFGWC is relatively faster than FGWC. However, the determination of exact, crisp values of the context variable is a hard task. Since this determination is quite important and affects the final clustering results so that it should be studied carefully.

Our contribution in this paper is the introduction of two novel fuzzy approaches for the determination of values of the context variable in CFGWC. The former named as CFGWC_F1 uses fuzzy clustering as a tool to determine the exact values of the context variable, and the later named as CFGWC_F2 calculate the values through type-2 fuzzy memberships. A numerical example is given to illustrate the uses of the proposed methods. The rest of the paper is structured as follows. Section 2 takes an overview of CFGWC algorithm and points out its limitations. Section 3 presents two novel algorithms CFGWC_F1 and CFGWC_F2. Section 4 shows a numerical example and the comparison of those algorithms. The conclusions and further works are summarized in Section 5.

## 2. CONTEXT FUZZY GEOGRAPHICALLY WEIGHTED CLUSTERING

### 2.1. Overview

Now, we summarize the principal ideas and details of CFGWC algorithm in [18]. Given a geo-demographic dataset of $N$ attributes $X = \{X_1,...,X_N\}$ in $r$ - dimension space ($X \in R^r$) with $X_k$ being the $k^{th}$ point. Supposed that missing data have been processed, the purpose of CFGWC is to classify the data into $C$ clusters, and $V_j$ is the center of $j^{th}$ cluster. A context variable in $Y \in X$ is defined as follows.

$$A : Y \to [0,1]$$
$$y_k \mapsto f_k = A(y_k).$$
(1)

$f_k$ is the representation for the level of relation of the $k^{th}$ point to the supposed context $Y$. There are some ways to define the relation between $f_k$ and the membership of $k^{th}$ point to the $j^{th}$ cluster, for instance, using the sum operator (2) or maximum operator (3).

$$\sum_{j=1}^{C} u_{kj} = f_k, \tag{2}$$

$$\max_{j=1}^{C} u_{kj} = f_k, \tag{3}$$

where $k = \overline{1, N}$. The basic objective function is,

$$J = \sum_{k=1}^{N} \sum_{j=1}^{C} u_{kj}^m \|X_k - V_j\|^2 \to \min. \tag{4}$$

where $m$ is the fuzziness and $u_{kj}$ is an element of the partition matrix $U$ below.

$$U(f) = \left\{ u_{kj} \in [0,1] : \sum_{j=1}^{C} u_{kj} = f_k, \forall k = \overline{1, N}, 0 < \sum_{k=1}^{N} u_{kj} < N, \forall j = \overline{1, C} \right\}. \tag{5}$$

Using the Lagranian method, the problem (4-5) is solved and details of the iteration scheme so-called CFGWC are shown as follows.

1. Initiate the matrix $U(t)$ at $t = 0$.

2. Re-calculate centers of each cluster according to equation (6).

$$V_j = \frac{\sum_{k=1}^{N} u_{kj}^m \times X_k}{\sum_{k=1}^{N} u_{kj}^m}, \quad j = \overline{1, C}. \tag{6}$$

3. Re-calculate matrix $U(t+1)$.

$$u_{kj} = \frac{f_k}{\sum_{i=1}^{C} \left( \frac{\|X_k - V_j\|}{\|X_k - V_i\|} \right)^{\frac{2}{m-1}}}, \quad k = \overline{1, N}, \quad j = \overline{1, C}. \tag{7}$$

4. Adjust the partition matrix by the SIM-PF model.

$$u_{kj}' = \alpha \times u_{kj} + \beta \times \frac{1}{A} \times \sum_{i=1}^{c} w_{ij} \times u_{ki}, \quad k = \overline{1, N}, \quad j = \overline{1, C}, \tag{8}$$

$$\alpha + \beta = 1, \tag{9}$$

$$w_{ij} = \frac{(pop_i \times pop_j)^b}{d_{ij}^a}, \tag{10}$$

where $u_{kj}'$ ($u_{kj}$) is the new (old) cluster membership of $k^{th}$ point to the $j^{th}$ cluster. Two parameters $\alpha$ and $\beta$ are scaling variables, and $A$ is a factor to scale the "sum" term to $f_k$ as in (5). $w_{ij}$ is the weight showing the influence of area $i$ to $j$. $pop_i$ ($pop_j$) is the population of area $i$ ($j$). $d_{ij}$ is the distance between those areas, and $a$ and $b$ are user definable parameters.

5. If the error of the partition matrix $\|U'(t+1) - U(t)\|$, defined through some analysis normal, is less than a given threshold $\varepsilon$ then the algorithm stops, else return to Step 2.

## 2.2. The limitation of CFGWC

In the article in [18], the authors used a random generator to create context values. As we can recognize in the description of CFGWC, it is hard to determine the exact value of $f_k$ ($k = \overline{1, N}$) to the supposed context $Y$ so that the quality of clustering outputs is not high as a result. For example, a person in the developed countries may assume 500,000 USD per year is "High" for the context "Income", but another in the developing countries can also state that 25,000 USD per year is "High". Misleading assessment of context values reduces the clustering quality of results, and therefore automatic determination of suitable values of the context variable should be done to handle this obstacle.

## 3. THE PROPOSED FUZZY APPROACHES

In this section, we present two fuzzy approaches to handle the limitation of CFGWC.

### 3.1. CFGWC_F1

The basic idea of CFGWC_F1 is using fuzzy clustering for the context variable and assigning the membership values of maximal context type to the context values. The number of clusters in this task is equal to that of the original problem. The reason for doing so is to handle the vagueness in the determination of exact, crisp values of the context variable. Fuzzy clustering, especially FCM algorithm is the most suitable tool to extract the knowledge behind an event or a context where the boundaries between clusters are unclear. Details of CFGWC_F1 are listed below.

1. Separate a subset of the original geo-demographic dataset containing the data of the supposed context $Y$ only.

2. Use FCM to divide the subset into $C$ clusters and get the partition matrix $U_C$.

3. For each $k = \overline{1, N}$:

a. Find the membership value of maximal context type in line $k^{th}$ of $U_C$.

b. Assign it to the context value of $X_k$

4. For all context values that have been calculated, we use the CFGWC algorithm in Section 2 to determine the final centers and membership values.

Since the membership values of maximal context type of the partition matrix $U_C$ reflects the maximal possibility of data points to given clusters, and the number of clusters in this task is equal to that of the original problem, thus those values can be used to orient the whole algorithm to the supposed context. In this case, their meanings are similar to those of the context variable.

### 3.2. CFGWC_F2

Now, we propose another way to determine the context values. Let us have a look at equation (1). The role of $Y$ is similar to that of the traditional fuzzy set if we re-written $Y$ as,

$$Y = \left\{ (y_k, f_k) \mid f_k \in [0,1], k = \overline{1, N} \right\}. \tag{11}$$

The limitations of the traditional fuzzy set were pointed out by Mendel in [8] including the definition of hard memberships so that fuzzy set cannot model some phenomena in real world. Such these sets cannot process some exceptional cases where the membership degrees are not the crisp values but the fuzzy ones instead. For example, the possibility to get tuberculosis disease of a patient concluded by a doctor is from 60 to 80 percents after examining all symptoms. Even if some modern medical machines are provided, the doctor cannot give an exact number of that possibility. This shows the fact that crisp membership values cannot model some situations in the real world and should be replaced with the fuzzy ones. Using traditional fuzzy sets often results in bad clustering quality since its uncertainties such as distance measure, fuzziness, center, prototype and initialization of prototype parameters can create imperfect representations of the pattern sets. For example, it is difficult to choose the suitable value for fuzziness. In case of pattern sets that contain clusters of different volume or density, it is possible that patterns staying on the left side of a cluster may contribute more for the other rather than this cluster. Similarly, how to choose a distance measure for fuzzy clustering is worth considering. Bad selection can yield undesirable clustering results for pattern sets that include noises. In order to handle the limitation of the traditional fuzzy set, in [8] Mendel suggested using the type-2 fuzzy set defined through the equation bellows.

$$\tilde{A} = \{(x, u, \mu_{\tilde{A}}(x, u)) \mid \forall x \in A, \forall u \in J_x \subseteq [0,1]\}. \quad (12)$$

The type-2 fuzzy set is a generalization of the traditional fuzzy set since we will get the traditional fuzzy set when there is no uncertainty in the third dimension. Based upon equation (12), equation (11) is re-written as,

$$\tilde{Y} = \{(y_k, f_k, \mu_{\tilde{A}}(y_k, f_k)) \mid \forall y_k \in Y, \forall f_k \in [0,1]\}, \quad (13)$$

$$f_k = \exp\left(\frac{-(y_k - \mu_Y)^2}{\sigma_Y^2}\right), \quad (14)$$

$$\mu_{\tilde{A}}(y_k, f_k) = \frac{1}{1 + e^{-f_k}}, \quad (15)$$

Where $\mu_Y$ and $\sigma_Y$ are the mean and the standard deviation of $Y$. Details of CFGWC_F2 are listed below.

1. For the supposed context $Y$, use equation (15) to calculate all context values.

2. For all context values that have been calculated, we use the CFGWC algorithm in Section 2 to determine the final centers and membership values.

### 3.3. Complexity

The time complexities of the context values calculation in both CFGWC_F1 and CFGWC_F2 are $O(N \times C)$ and $O(N)$, respectively.

## 4. A NUMERICAL EXAMPLE

We have implemented the proposed algorithms (CFGWC_F1 and CFGWC_F2) in addition to CFGWC in [18] in C programming language and executed them on a PC with configuration: Intel Pentium Dual Core 1.80 GHz, 1GB RAM. The objective of experiments is to verify the impacts of the context generation methods in CFGWC_F1 and CFGWC_F2 to the clustering quality of outputted results in comparison with the random context generation method in CFGWC in [18]. In the other words, we aim to answer whether or not the clustering qualities of CFGWC_F1 and CFGWC_F2 are better than that of CFGWC. The experimental dataset was taken from the articles in[16], [17] and a small part of it is described in Table 1. Parameters of CFGWC_F1 and CFGWC_F2 are set up similar to those of CFGWC as in [18]. Experimental results are listed step-by-step to illustrate the activities of the proposed algorithm. In Table 1, the chosen context variable is "Income". We would like to divide the dataset above into three clusters according to the context variable, which are "Low income", "High income" and "Medium income".

Table 1. The statistics of geo-demographic characteristics

| Name  | Occupation | Income | Age | Gender | Raise |
|-------|------------|--------|-----|--------|-------|
| Marry | Student    | 28,000 | 15  | Female | 4     |
| Tom   | Doctor     | 40,000 | 32  | Male   | 2     |
| David | Doctor     | 35,100 | 27  | Male   | 6     |
| Kim   | Singer     | 65,000 | 19  | Female | 1     |
| Jenny | Student    | 20,000 | 18  | Female | 3     |
| Julia | Singer     | 52,520 | 23  | Male   | 6     |
| Xiao  | Student    | 21,000 | 31  | Male   | 3     |
| Luka  | Doctor     | 75,000 | 42  | Female | 2     |

Now we illustrate the activities of CFGWC_F1. The subset containing the data of the supposed context is:

$$Y = \{28,000; 40,000; 35,100; 65,000; 20,000; 52,520; 21,000; 75,000\}. \quad (16)$$

Use FCM to divide $Y$ into 3 groups, we receive the membership values $U_C$.

$$U_C = \begin{pmatrix} 0.830213 & 0.013943 & 0.155844 \\ 0.000256 & 0.000091 & 0.999653 \\ 0.145173 & 0.019431 & 0.835396 \\ 0.008823 & 0.965323 & 0.025853 \\ 0.979944 & 0.002928 & 0.017128 \\ 0.098034 & 0.319610 & 0.582355 \\ 0.991251 & 0.001213 & 0.007536 \\ 0.012425 & 0.959367 & 0.028209 \end{pmatrix}. \quad (17)$$

According to equation (17), we have a preliminary classification of all users according to the context "Income". For example, in line $8^{th}$ of $U_C$, the second value 0.9599367 is the largest among all. Thus, the income of user "Luka" is considered as "High". Similarly, in line $5^{th}$ of $U_C$, the first value 0.979944 is the largest among all. Thus, the income of user "Jenny" is considered as "Low". Now, we take the membership values of "High income" as the context values and apply the CFGWC algorithm for them. The final membership values and centers are:

$$U^1 = \begin{pmatrix} 0.005391 & 0.003337 & 0.005303 \\ 0.000021 & 0.000037 & 0.000036 \\ 0.007412 & 0.004669 & 0.007410 \\ 0.364774 & 0.211715 & 0.378353 \\ 0.001169 & 0.000704 & 0.001136 \\ 0.088755 & 0.135399 & 0.121199 \\ 0.000465 & 0.000306 & 0.000455 \\ 0.349459 & 0.243099 & 0.352343 \end{pmatrix}, \quad (18)$$

$$V^1 = \begin{pmatrix} 2.53 & 68853.72 & 29.72 & 1.05 & 1.70 \\ 2.52 & 69129.90 & 30.13 & 1.05 & 1.71 \\ 2.52 & 68863.76 & 30.12 & 1.06 & 1.79 \end{pmatrix}. \quad (19)$$

Figure 1. describes the distribution of data points resulted by CFGWC_F1. Now we illustrate the activities of CFGWC_F2. Firstly, we calculate the mean and the standard deviation of $Y$ as follows.

$$\mu_Y = \frac{\sum_{k=1}^{8} X_{kY}}{8} = 42077.5, \quad (20)$$

$$\sigma_Y = \sqrt{\frac{\sum_{k=1}^{8}(X_{ky} - \mu_Y)^2}{8}} = 19043.47. \quad (21)$$

Use the formulas in equations (14-15), we calculate the context values.

$$\mu_{\tilde{A}}^T = \begin{pmatrix} 0.64 & 0.73 & 0.7 & 0.56 & 0.56 & 0.68 & 0.57 & 0.51 \end{pmatrix}. \quad (22)$$

Use those context values in (22) for the CFGWC algorithm and get the final membership values and centers are.

$$U^2 = \begin{pmatrix} 0.26 & 0.24 & 0.14 \\ 0.46 & 0.24 & 0.03 \\ 0.3 & 0.26 & 0.14 \\ 0.19 & 0.2 & 0.17 \\ 0.21 & 0.2 & 0.15 \\ 0.2 & 0.23 & 0.25 \\ 0.22 & 0.21 & 0.14 \\ 0.17 & 0.18 & 0.16 \end{pmatrix}, \quad (23)$$

$$V^2 = \begin{pmatrix} 1.89 & 40671.23 & 25.7 & 1.58 & 3.58 \\ 1.97 & 42088.29 & 26.2 & 1.62 & 3.68 \\ 2.03 & 43297.50 & 27.1 & 1.68 & 3.81 \end{pmatrix}. \quad (24)$$

Figure 2 describes the distribution of data points resulted by CFGWC_F2.

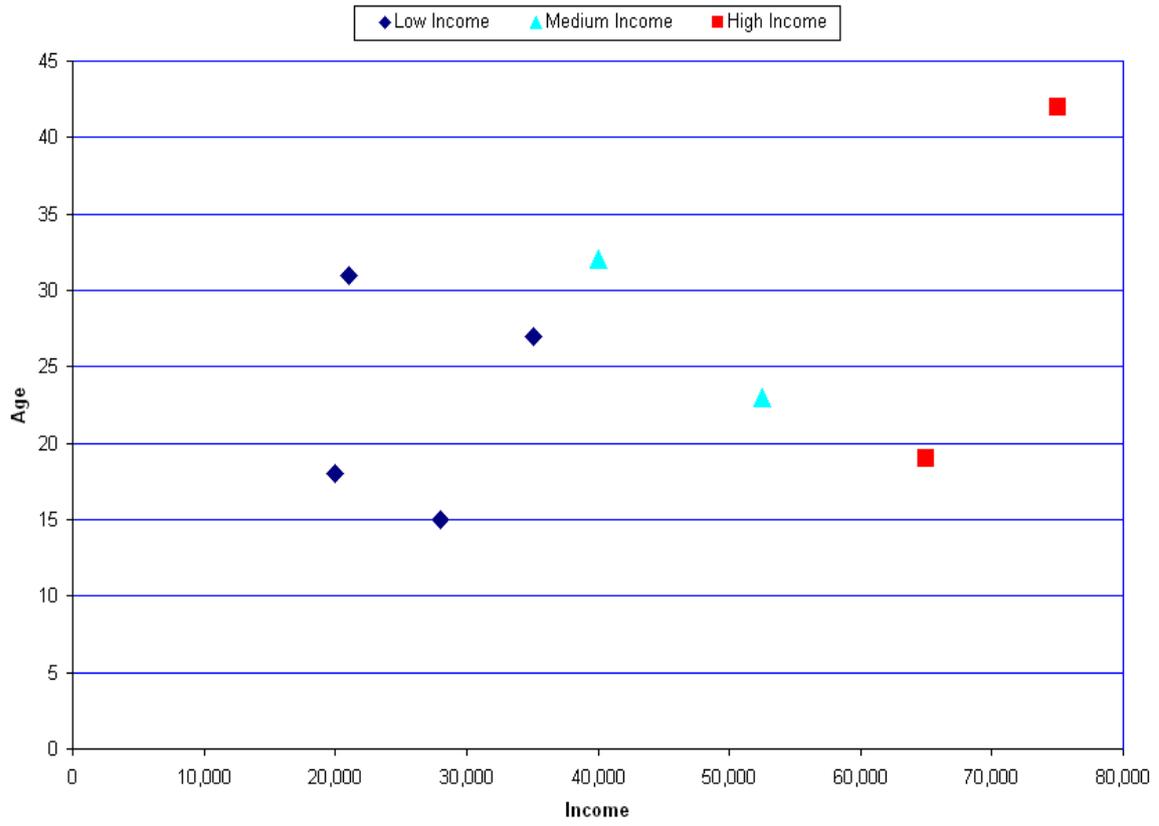

Figure 1. The results of CFGWC_F1

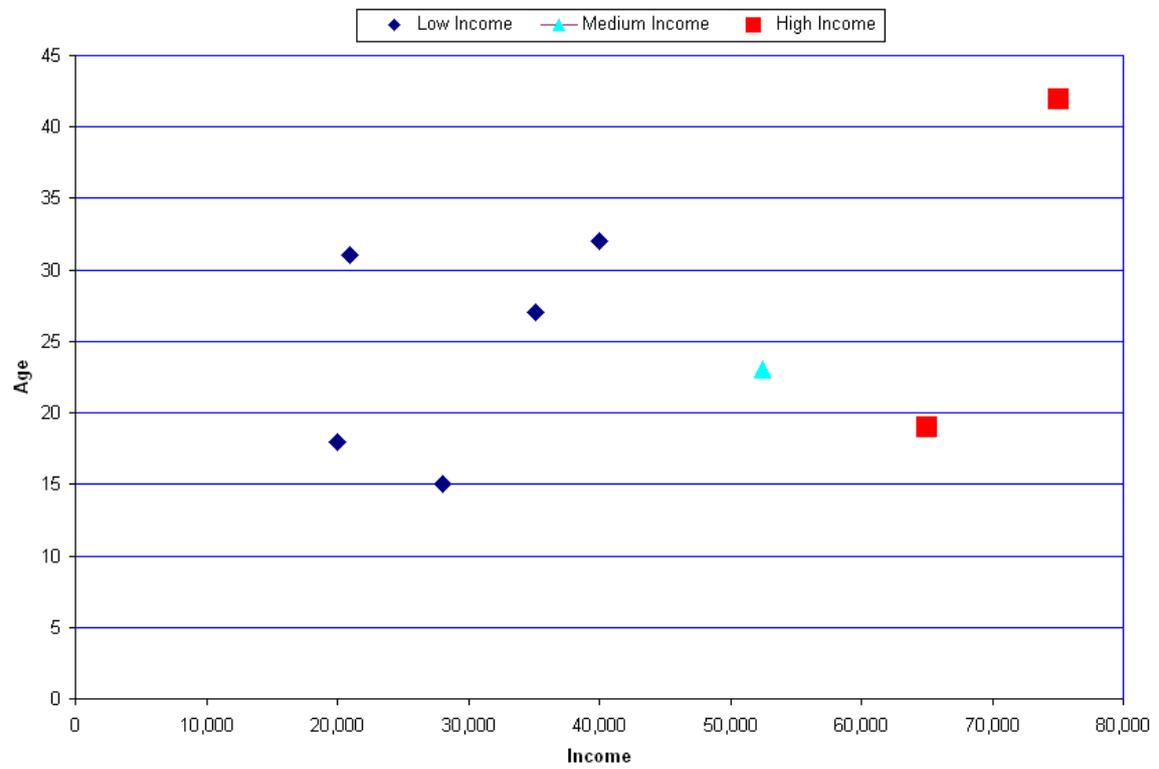

Figure 2. The results of CFGWC_F2

In order to investigate the effects of using various context generation methods to CFGWC, we make a comparison of the clustering quality between these algorithms using the validity function of fuzzy clustering for spatial data namely IFV in [4]. This index was shown to be robust and stable when clustering spatial data. The definition of this index is characterized below.

$$IFV = \frac{1}{C} \sum_{j=1}^{C} \left\{ \frac{1}{N} \sum_{k=1}^{N} u_{kj}^2 \left( \log_2 C - \frac{1}{N} \sum_{k=1}^{N} \log_2 u_{kj} \right)^2 \right\} \times \frac{SD_{\max}}{\overline{\sigma_D}}. \quad (25)$$

The maximal distance between centers is:

$$SD_{\max} = \max_{k \neq j} \left\| V_k - V_j \right\|^2. \quad (26)$$

The even deviation between each object and the cluster centre is:

$$\overline{\sigma_D} = \frac{1}{C} \sum_{j=1}^{C} \left( \frac{1}{N} \sum_{k=1}^{N} \left\| X_k - V_j \right\|^2 \right). \quad (27)$$

When $IFV \to \max$, the value of $IFV$ is said to yield the most optimal of the dataset.

From equations (18, 19, 23, 24), we calculate IFV values of CFGWC_F1 and CFGWC_F2.

$$IFV^{CFGWC\_F1} = 8.535490, \quad (28)$$

$$IFV^{CFGWC\_F2} = 8.321658. \quad (29)$$

Besides, we also calculate IFV value of CFGWC and receive the result in equation (30).

$$IFV^{CFGWC} = 7.553624. \quad (30)$$

From equations (28-30), we recognize that the clustering qualities of CFGWC_F1 and CFGWC_F2 are better than that of CFGWC. Additionally, CFGWC_F1 is better than CFGWC_F2. The distributions of data points in both methods are shown in Figure 1 and Figure 2

## 5. CONCLUSIONS

In this paper, we introduced two novel fuzzy approaches to determine suitable context values for fuzzy geographically weighted clustering. The former used fuzzy clustering as a tool to determine the exact values of the context variable, and the later calculated the values through type-2 fuzzy memberships. A numerical example was given to illustrate the uses of the proposed methods. The results showed that the clustering qualities of the proposed methods are better than that of the relevant one. Further works of this paper will investigate multiple contexts and their suitable orders in clustering algorithms.

### ACKNOWLEDGEMENTS

This work is sponsored by a VNU project under contract No. QG.13.01

**Authors**

Msc**. Nguyen Van Minh** is a lecturer at the Faculty of Science, Hanoi University of Natural Resources and Environment. His research interests include Fuzzy Clustering, Recommender Systems, Soft Computing and Statistic. Office address: Faculty of Science, Hanoi University of Natural Resources and Environment, 41A, Phu Dien, Bac Tu Liem, Hanoi, Vietnam. Email: nvminh@hunre.edu.vn

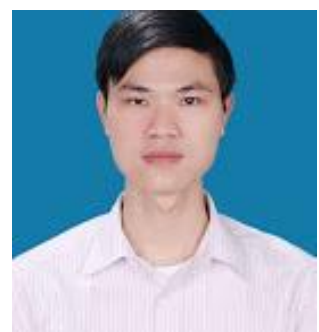

Dr. **Le Hoang Son** obtained the PhD degree on Mathematics – Informatics at VNU University of Science, Vietnam National University. Currently, he is a researcher at the Center for High Performance Computing, VNU University of Science, Vietnam National University. His major field includes Soft Computing, Fuzzy Clustering, Recommender Systems, Geographic Information Systems and Particle Swarm Optimization. He is a member of IACSIT and also an associate editor of the International Journal of Engineering and Technology (IJET). He also served as a reviewer for various international journals and conferences such as PACIS 2010, ICMET 2011, ICCTD 2011, KSE 2013, BAFI 2014, NICS 2014, ACIIDS 2015, International Journal of Computer and Electrical Engineering, Imaging Science Journal, International Journal of Intelligent Systems Technologies and Applications, IEEE Transactions on Fuzzy Systems, 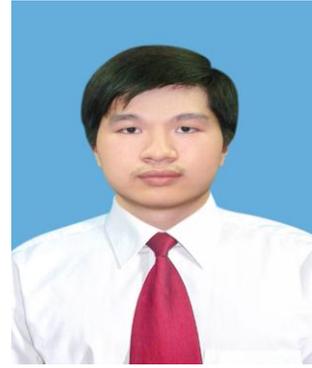 Expert Systems with Applications, and International Journal of Electrical Power and Energy Systems. Dr. Son has got many publications in prestigious journals and undertaken some major research projects of Vietnam and international joint projects. Office address: VNU University of Science, Vietnam National University, 334 Nguyen Trai, Thanh Xuan, Hanoi, Vietnam. Email: sonlh@vnu.edu.vn